# Analyzing Transformer Models and Knowledge Distillation Approaches for Image Captioning on Edge AI

Wing Man Casca Kwok
Independent researcher
San Jose, US
kwok.wi@northeastern.edu

Yip Chiu Tung
Independent researcher
Hong Kong
tungchris@gmail.com

Kunal Bhagchandani
Independent researcher
San Jose, US
kunal.bhagchandani10@gmail.com

*Abstract*— **Edge computing decentralizes processing power to network edge, enabling real-time AI-driven decision-making in IoT applications. In industrial automation such as robotics and rugged edge AI, real-time perception and intelligence are critical for autonomous operations. Deploying transformer-based image captioning models at the edge can enhance machine perception, improve scene understanding for autonomous robots, and aid in industrial inspection.**

**However, these edge or IoT devices are often constrained in computational resources for physical agility, yet they have strict response time requirements. Traditional deep learning models can be too large and computationally demanding for these devices.**

**In this research, we present findings of transformer-based models for image captioning that operate effectively on edge devices. By evaluating resource-effective transformer models and applying knowledge distillation techniques, we demonstrate inference can be accelerated on resource-constrained devices while maintaining model performance using these techniques.**

## I. Introduction

Recent developments in multi-modal deep learning have opened up exciting use of generative image systems, but their impact isn't limited to text-to-image or image-to-text generation. They are transforming industries that rely on image analytics to extract contextual information.

This includes fields like human-machine interaction, where traditional object detection systems may not be adequate to richly describe a scene.

Multi-modal systems can also enhance unimodal classifier accuracy by incorporating additional modalities. For instance, research work in medical imaging has shown that training an image classifier with integrated scans and textual medical records can improves diagnostic accuracy[19].

Another common usage is image annotation. Traditionally, image annotation relies heavily on human labor. Automated image captioning systems could address this challenge by providing a solution for generating image descriptions.

These image captioning applications, when used at network edge, need to be adaptable to diverse computational environments such as portable machine and IoT devices mentioned above. While the current state-of-the-art deep learning models for both vision and natural language processing (NLP) tasks are transformers[1], typical transformer models are large in parameter size, usually require cloud or significant compute resources for inference.

To address this, our work contributes to advancing the field of edge intelligence by analyzing and identifying resource-effective transformer encoder-decoder models, their variants and distilled variants, architecture, to uncover models and methodologies, that can be computationally effective to run on edge devices.

Transformer-based image captioning systems are generally composed of an encoder and a decoder. An encoder processes visual information and extracts features from images, while a decoder translates those features into a textual information. During inference, a decoder operates auto-progressively, predicting one word at a time by using both the previously generated words and the visual features from the encoder.

## II. Related work

Transformer has emerged as the current state-of-the-art in natural language processing. Its fully attention-based mechanism has eliminated the requirement of recurrence, the process which requires sequential processing. Without this limitation, data is allowed to be processed more efficiently through parallelization. In addition, transformer's self-attention mechanism enables each block in the input sequence to attend to every other element, thus it can effectively capture semantic relevance across the entire sequence regardless of distance.

Building on this success, Vision Transformer (ViT)[2] was introduced in 2021 as the first transformer-based architecture with primary use on computer vision tasks. ViT processes images by dividing them into patches, treating these patches as tokens, in a similar fashion of how words are processed into tokens in NLP tasks. The use of ViT allows for the capture of both visual and spatial relationships across image patches. The patches are also self-attended to predict inter-patch relevance, thus ViT provides a comprehensive understanding of an image representation, and the approach has gained popularity.

The transformer architecture later evolved to introduce cross-attention between image and text embedding. The architecture predicts textual information or caption from an image by also considering the relationships between image patches and contextual information. Cross-attention is the key reason we chose transformer architecture in our work.

ViT is an encoder architecture. In its vanilla form, it is composed of 12 layers. It is known to require substantial datasets for effective training. Being the first transformer architecture applied to computer vision, ViT was trained on 300 million images and was found not generalizing well when trained on insufficient amounts of data.

Hugo et al. introduced Data-efficient image transformers (DeiT)[3] to enhance the efficiency of dataset usage when training vision transformers. They employed training strategies such as strong data augmentation, and experimented to find optimum regularization strategies. The model was trained on the Imagenet1k dataset composed of 1,281,167 training images. Their approach achieved higher top-1 accuracy[15] when compared with ViT, despite using less than half of the training images that ViT was trained on.

To make these models lightweight and low latency for the use on compute-light and edge devices, a common approach is to leverage model compression methodologies such as pruning, quantization, and knowledge distillation to reduce the number of model parameters, thereby accelerating inference. Knowledge distillation[4] is an approach, where a compact model, commonly referred to as student, is trained to mimic the predictions of an accurate teacher model. This method can build a compact model that maintains accuracy and performance of the teacher model, making it suitable with compute-light edge devices.

DeiT in its distilled form, employs a unique distillation procedure by adding a distillation token to the sequence during training. This approach has been found outperforming common distillation methods, which typically focus on learning the teacher's class probabilities, known as soft labels.

BERT(Bidirectional Encoder Representations from Transformers)[5] has been established as the state-of-art transformer model in NLP. Based on the transformer architecture, BERT consider context of a word from both directions in a sequence, and the approach has set new benchmarks of accuracy in natural language processing tasks.

TinyBERT[6] is a distilled BERT model composed of only 4 layers. It is 7.5x smaller, maintaining more than 96.8% prediction performance and being 9.4x faster while on inference when compared with its $\text{BERT}_{\text{BASE}}$ teacher. In Xiaoqi Jiao et al.'s work, they introduced a knowledge distillation method which is purpose-built for transformer networks. Unlike common distillation approaches where a student model learns from the teacher's soft label, the TinyBERT student learns from the outputs of the teacher's multi-head attention layer, prediction layer and embedding layers.

Another distilled BERT approach introduced in Iulia et al.'s work[7] demonstrates that by pre-training a student network before it goes through distillation, making it well-read to prepare to take the full knowledge of the teacher, can outperform models that only undergo pre-training and fine-tuning, or those that are distilled without pre-training. Their small version, $\text{Transformer}_{\text{TINY}}$, is 77 times smaller than its $\text{BERT}_{\text{LARGE}}$ teacher while 65 times faster. Their distillation process takes the common approach on training on teacher's soft labels.



| **Image** |
|---|
| COCO_train2014_000000540135.jpg |
| **Caption** |
| a woman watching a dog jump up for a frisbee |
| A dog in mid air catching a frisbee on a field. |
| A dog and its trainer play with Frisbees on a lawn. |
| A dog that is catching a frisbee in the air. |
| A dog is cathing a Frisbee a woman is holding an orange Frisbee in hand |

Fig. 1. An example of the COCO 2014 training dataset, where each image is associated with 5 captions.

## III. Method

We utilized the COCO 2014 dataset[10] for training and validation, and COCO 2017 dataset for inference. The COCO 2014 dataset consists of 82,783 training, 40,504 validation, and 40,775 testing samples. Each image is associated with 5 captions (Fig. 1). We used 18,000 training and 1,000 validation samples to obtain the results.

In our work, we employed ViT, DeiT, DeiT Tiny and distilled DeiT tiny as the image encoders, while utilizing BERT, TinyBERT and Transformer TINY, (Transformer TINY is named by Iulia et al.'s work for their pre-trained distilled BERT variant), as text decoders.

One of the key parts of the encoder-decoder architecture is the cross-attention head, which takes the key and value matrices from the encoder and the query matrix from the decoder embeddings to compute the attention between image and text.

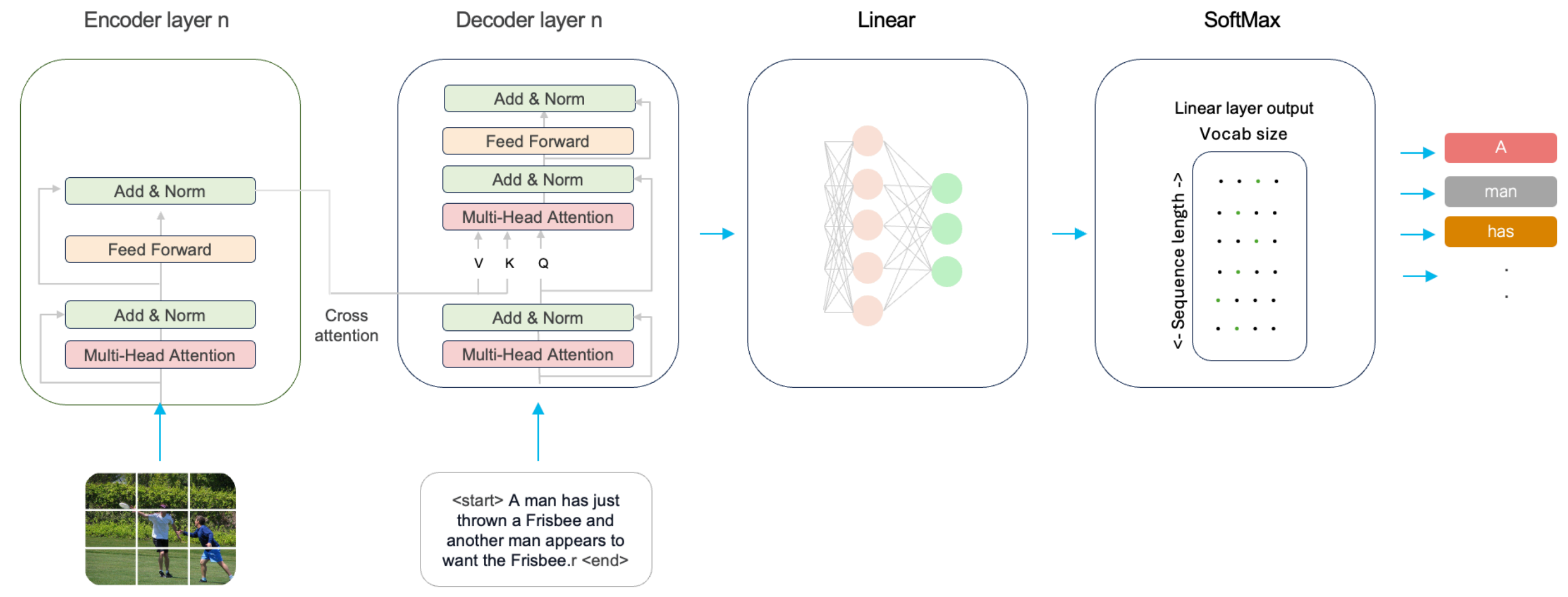

Fig. 2. Image captioning using a transformer encoder-decoder architecture. An image encoder uses self-attention mechanisms to uncover inter-patch spatial relationships. The image embeddings are then cross-attended with text embeddings at a decoder to generate captions.

At each training step, the model predicts the sequence of words by comparing it with the caption ground truth. (Fig. 2)

We evaluated the performance outcomes of these mixed encoder-decoder configurations (TABLE I. ) in relation to the combined model size (TABLE II. ). We also assessed inference time improvements between the teacher models, their variants and students, to observe for the effectiveness of knowledge distillation in image captioning tasks.

The networks were fine-tuned from their corresponding encoder and decoder pre-trainings using a batch size 16 and a learning rate 5e-5. They were trained for 18 epochs with early stopping. The images used were 224 x 224 pixels normalized with ImageNet mean and standard deviation. To facilitate text generation, we used a vocabulary size of 30552. The models were fine-tuned on a Nvidia A100 GPU platform with 40GB of GPU RAM.

TABLE I. COMPARISON OF MODEL STRUCTURE AND NUMBER OF PARAMETERS

| Model | Embedding dimension | # layers | Parameters |
|---|---|---|---|
| DeiT | 768 | 12 | 86M |
| ViT | 768 | 12 | 86M |
| DeiT tiny | 192 | 12 | 5M |
| Distilled DeiT tiny | 192 | 12 | 6M |
| BERT | 768 | 12 | 86M |
| TinyBERT | 312 | 4 | 14M |
| Transformer $_{TINY}$ | 128 | 2 | 4M |

To measure model performance, we used ROUGE(Recall-Oriented Understudy for Gisting Evaluation) and BLEU(Bilingual Evaluation Understudy) as the evaluation metrics. These metrics measure the degree of similarity between the generated text and the ground truth caption.

ROUGE [13] measures the overlapping of n consecutive items, known as n-gram, between the generated text and the reference text. ROUGE-1 measures the overlap of unigrams (single words) which assess word-level similarity. ROUGE-2 measures the overlap of bigrams (two-word sequences) to capture the relationship between consecutive words. ROUGE-L measures the longest sequence of words that appear in both the generated and reference texts to measure structural similarity. ROUGE-Lsum averages ROUGE-L scores for each individual sentence, making it suitable for tasks where sentence level information extraction is desired.

BLEU [14] measures the precision of unigrams, bigrams up to 4-grams.

TABLE II. COMPARISON OF COMBINED MODEL PARAMETERS

| Encoder | Decoder | Parameters |
|---|---|---|
| DeiT | BERT | 224,270,394 |
| ViT | BERT | 224,270,394 |
| Distilled DeiT tiny | TinyBERT | 21,568,458 |
| DeiT tiny | TinyBERT | 21,568,074 |

| | | | | | |
|---|---|---|---|---|---|
| Distilled DeiT tiny | Transformer TINY | 10,135,866 | DeiT tiny | Transformer TINY | 10,135,482 |

TABLE III. EVALUATION RESULTS WITH 18,000 TRAINING SAMPLES

| Encoder | Decoder | ROUGE -1 | ROUGE -2 | ROUGE -3 | ROUGE -L sum | BLEU |
|---|---|---|---|---|---|---|
| ViT | BERT | 0.3854 | 0.1338 | 0.3510 | 0.3513 | 0.0821 |
| DeiT | BERT | 0.3790 | 0.1284 | 0.3446 | 0.3446 | 0.0784 |
| DeiT tiny | TinyBERT | 0.3553 | 0.1143 | 0.3264 | 0.3267 | 0.0726 |
| DeiT tiny | Transformer TINY | 0.3404 | 0.1046 | 0.3161 | 0.3161 | 0.0603 |
| Distilled DeiT tiny | TinyBERT | 0.3596 | 0.1153 | 0.3298 | 0.3297 | 0.0724 |
| Distilled DeiT tiny | Transformer TINY | 0.3509 | 0.1127 | 0.3253 | 0.3254 | 0.0686 |

# IV. RESULTS

## 1. *Evaluation performance*

ViT + BERT and DeiT + BERT achieved the highest evaluation performance as expected. The most accurate variant, Distilled DeiT tiny + TinyBERT, attained 95% of the ROUGE-1 performance of their DeiT + BERT teacher (TABLE. III), while achieving 90% reduction in parameters with a combined parameter counts of 21M. DeiT tiny + Transformer TINY, being the most compact combined models, achieved 90% of the teachers' performance. They are smaller than distilled DeiT tiny + Transformer TINY because the latter introduces additional distillation tokens, which slightly increased the number of parameters.

## 2. *Inference time and compute resources efficiency improvement*

Regarding inference time improvement, on a platform with 1 vCPU and 4G RAM (compute resources configured via Windows Subsystem for Linux (WSL2), on a quad-core Intel i7@ 3GHz CPU with 16G RAM on a Lenovo Thinkpad), Distilled DeiT tiny + TinyBERT achieved a 5.4x speedup in processing time (TABLE. IV). Notably, DeiT tiny + Transformer TINY achieved the lowest inference time, with a 12.5x speedup, taking only 193ms for inference.

For compute resources improvement, the actual memory consumed when running the inference of Distilled DeiT tiny + TinyBERT is 850MB, which is a 58% saving compared to their teacher models. For storage spaces required for the model weights, the most compact models DeiT tiny + Transformer TINY occupy only 41MB storage, which is a 95% reduction from the teacher models.

## 3. *Effectiveness of layer-freezing when fine-tuning*

We attempted to save fine-tuning time and cost by freezing the weights of earlier layers and only training the last two layers of the encoders and decoders. Teacher models like ViT + BERT were able to converge stably. However, for some of the variants, the trainings did not consistently converge due to an insufficiency of parameters. We found that when all layers were unfrozen, their trainings maintained stable convergence, also resulting better evaluation performance.

## 4. *Dataset size*

We attempted to fine-tune the models with 9,000 training samples as the teacher models can produce satisfactory performance. However, for the compact models, we found that the captions generated did not match quite well with the scene of an image. By increasing training samples to 18,000, we observed good improvement.

TABLE IV. COMPARISON OF INFERENCE TIME, ACTUAL MEMORY USAGE DURING INFERENCE AND MODEL WEIGHT STORAGE SIZE

| Encoder | Decoder | Inference Time (s) | Memory Usage | Model Weight(MB) |
|---|---|---|---|---|
| ViT | BERT | 2.4155 | ~2GB | 897 |
| DeiT | BERT | 2.4039 | ~2GB | 897 |
| DeiT tiny | TinyBERT | 0.4373 | ~900MB | 86 |
| DeiT tiny | Transformer TINY | 0.1929 | ~780MB | 41 |
| Distilled DeiT tiny | TinyBERT | 0.4439 | ~850MB | 86 |
| Distilled DeiT tiny | Transformer $_{TINY}$ | 0.2535 | ~800MB | 41 |

TABLE V. GENERATED CAPTION BY THE TRANSFORMER ENCODERS AND DECODERS



| Encoder | Decoder | Generated caption |
|---|---|---|
| ViT | BERT | three men in suits standing next to each other. |
| DeiT | BERT | two men in suits and ties posing for a picture. |
| DeiT tiny | TinyBERT | a man and a woman standing next to each other. |
| | Transformer $_{TINY}$ | a group of men standing around a table. |
| Distilled DeiT tiny | TinyBERT | a man and a woman are holding a glass of wine. |
| | Transformer $_{TINY}$ | a couple of men standing next to each other. |

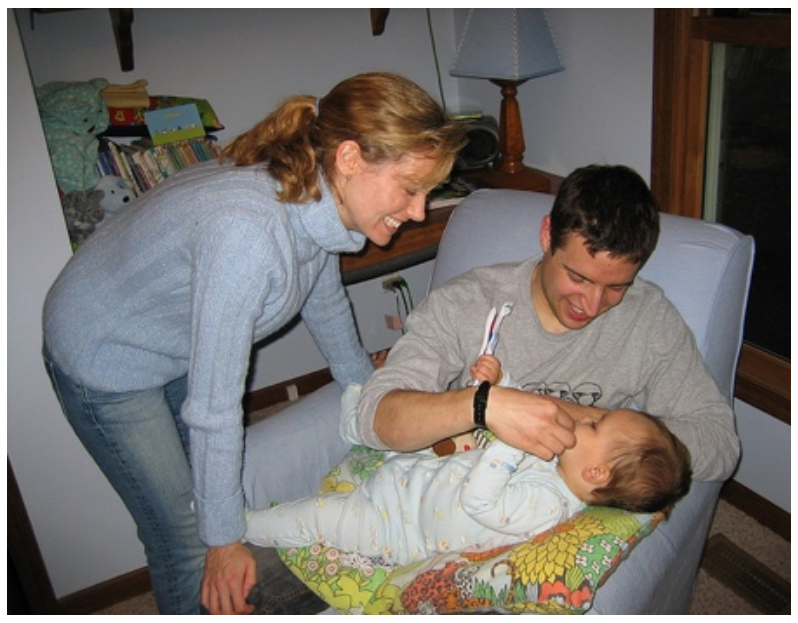

| Encoder | Decoder | Generated caption |
|---|---|---|
| ViT | BERT | a group of people are cutting a cake. |
| DeiT | BERT | a man and a woman sitting in a chair with a baby. |
| DeiT tiny | TinyBERT | two men sitting at a table with a baby. |
| | Transformer $_{TINY}$ | two men are playing a game of wii. |
| Distilled DeiT tiny | TinyBERT | a man and a woman are sitting in a chair with a baby. |
| | Transformer $_{TINY}$ | a man and a woman sitting on a couch. |

### 5. *Generated captions*

Table V provides a comparison of captions generated by each encoder-decoder combination. It shows that the teacher models offer richer context, while the student models can provide basic description that align with the scenes.

## V. CONCLUSIONS

In this research, we have demonstrated the effectiveness of integrating vision and language transformers that use cross-attention mechanisms in an encoder-decoder architecture, along with the strategies that enable computational efficiency for running image captioning systems on edge devices.

Through exploration with various encoding and decoding model configuration, the respective knowledge distillation methodologies, we assessed the performance outcomes of these models with their variants and distilled variants. These variants have achieved a 90 - 95% reduction in parameters size, while maintaining 90 - 95% of the ROUGE-1 performance compared to their teacher models.

These compact models enable efficient computation on edge devices, requiring less than 1GB of memory and under 100MB for storing training weights, while delivering up to 12.5x improvement in inference speed with a single vCPU.

Owing to the reduced size in parameters of these variants, more training data is required to ensure comparable performance.

This study demonstrates that transformer-based architectures can deliver effective image captioning solutions in compute-light environments for edge intelligence. For future work, we would further explore optimizations strategies of these models for the application on different edge domains.